\documentclass{article}

\usepackage{arxiv}

\usepackage[utf8]{inputenc} 
\usepackage[T1]{fontenc}    
\usepackage{hyperref}       
\usepackage{url}            
\usepackage{booktabs}       
\usepackage{amsfonts}       
\usepackage{nicefrac}       
\usepackage{microtype}      
\usepackage{lipsum}
\usepackage{graphicx}
\usepackage{float}
\graphicspath{ {./images/} }
\usepackage{amsmath}
\usepackage{graphicx} 

\title{UltraUNet: Real-Time Ultrasound Tongue Segmentation for Diverse Linguistic and Imaging Conditions\thanks{This is a preprint submitted to arXiv.}}

\author{
 Alisher Myrgyyassov \\
  Biomedical Engineering Department\\
  Hong Kong Polytechnic University\\
  Hong Kong, China  \\
  \texttt{alisher2.myrgyyassov@connect.polyu.hk} \\
  \And
 Zhen Song \\
  Biomedical Engineering Department\\
  Hong Kong Polytechnic University\\
  Hong Kong, China  \\
  \texttt{zhen0212.song@connect.polyu.hk} \\
  \And
 Yu Sun \\
  Biomedical Engineering Department\\
  Hong Kong Polytechnic University\\
  Hong Kong, China  \\
  \texttt{stefanie.sun@polyu.edu.hk} \\
  \And
 Bruce Xiao Wang \\
  English and Communication Department\\
  Hong Kong Polytechnic University\\
  Hong Kong, China  \\
  \texttt{brucex.wang@polyu.edu.hk} \\
  \And
 Min Ney Wong \\
  Department of Chinese and Bilingual Studies\\
  Hong Kong Polytechnic University\\
  Hong Kong, China  \\
  \texttt{min.wong@polyu.edu.hk} \\
  \And
 Yongping Zheng\\
  Department of Biomedical Engineering\\
  Research Institute for Smart Ageing\\
  Hong Kong Polytechnic University\\
  Hong Kong, China  \\
  \texttt{yongping.zheng@polyu.edu.hk} \\
}

\begin{document}
\maketitle
\begin{abstract}
Ultrasound tongue imaging (UTI) provides a non-invasive, cost-effective modality for investigating speech articulation, speech motor control, and speech-related disorders. However, real-time tongue contour segmentation remains a significant challenge due to the inherently low signal-to-noise ratio, variability in imaging conditions, and computational demands of real-time performance. In this study, we proposed UltraUNet, a lightweight and efficient encoder-decoder architecture specifically optimized for real-time segmentation of tongue contours in ultrasound images. UltraUNet introduces several domain-informed innovations, including lightweight Squeeze-and-Excitation blocks for channel-wise feature recalibration in deeper layers, Group Normalization for enhanced stability in small-batch training, and summation-based skip connections to minimize memory and computational overhead. These architectural refinements enabled UltraUNet to achieve a high segmentation accuracy while maintaining an exceptional processing speed of 250 frames per second, making it suitable for real-time clinical workflows. UltraUNet integrates ultrasound-specific augmentation techniques, including denoising and blur simulation using point spread function. Additionally, we annotated UTI images from 8 different datasets with various imaging conditions. Comprehensive evaluations demonstrated the model’s robustness and precision, with superior segmentation metrics on single-dataset testing (Dice = 0.855, MSD = 0.993px) compared to established architectures. Furthermore, cross-dataset testing on 7 unseen datasets with 1 train dataset revealed UltraUNet’s generalization capabilities and high accuracy, achieving average Dice Scores of 0.734 and 0.761, respectively, in Experiments 1 and 2. The proposed framework offers a competitive solution for time-critical applications in speech research, speech motor disorder analysis, and clinical diagnostics, with real-time performance in tongue functional analysis in diverse medical and research settings.
\end{abstract}

\section{Introduction}
Tongue movement patterns are a rich source of clinically important information, offering insights into speech articulation, speech motor control, and speech disorders. Accurate analysis of these patterns is essential in understanding both normal and pathological speech production \cite{RN1}. Ultrasound imaging has become a widely adopted tool in speech research due to its ability to visualize tongue movements dynamically in real-time, making it an invaluable resource for studying articulatory processes and diagnosing speech-related conditions \cite{RN4, RN3, RN2}. Additionally, ultrasound imaging is a non-invasive, safe, and cost-effective means of capturing tongue motion \cite{RN3}, making it a widely accessible alternative to modalities like electromagnetic articulography, Magnetic Resonance Imaging or X-rays, which involve higher costs or radiation exposure. Previous studies have also demonstrated its high repeatability and reproducibility, suitable for adoption in clinical and research settings \cite{RN8, RN7, RN6, RN5}.
Extracting tongue contour data is an essential step in retrieving the most clinically relevant information from ultrasound images \cite{RN9}. The extracted contours are necessary for the quantification of tongue shape, movement trajectories, and interaction with other articulatory structures. These data can support not only phonetic studies, where articulatory patterns are analyzed but also speech therapy applications. In speech therapy, real-time tongue contour extraction in B-mode ultrasound is particularly transformative. By isolating and analyzing the tongue's contours from ultrasound images, clinicians can utilize visual biofeedback to provide immediate and effective guidance to patients with speech disorders, facilitating targeted interventions in speech therapy \cite{RN8}. This facilitates speech motor control training, helping patients achieve more accurate tongue configuration during speech exercises. Ultrasound imaging feedback has been successfully applied in various clinical settings, particularly to address speech-related conditions \cite{RN10, RN11, RN12}. These applications include its use in treating individuals with persistent speech impairments, such as residual articulation errors \cite{RN8, RN7, RN5}. Ultrasound tongue imaging (UTI) feedback have demonstrated promising potential in speech therapy, particularly in cases where other approaches were not sufficiently effective, leaving unresolved speech impairments after treatment \cite{RN12}. 

\begin{figure}
    \centering
    \includegraphics[width=0.7\linewidth]{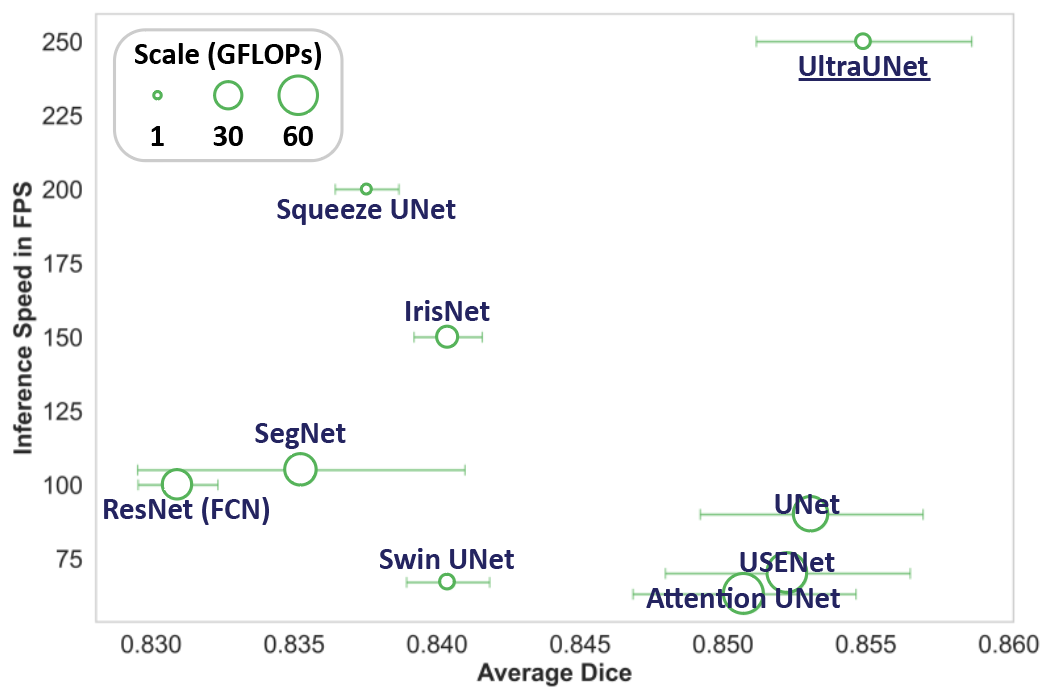}
    \caption{Dice vs Inference Speed (in FPS) plot for the Single-Dataset Evaluation experiment (Section 4.1), with the size of each point proportional to the number of Floating-Point Operations (FLOPs) for each model. Error bars are proportional to the standard deviation across all trials.}
    \label{fig:placeholder}
\end{figure}

To further enhance the efficiency of UTI-based therapy and research, the automation of tongue contour extraction was widely adopted. However, numerous challenges persist, particularly in achieving accurate and robust segmentation across varying imaging and linguistic conditions. For instance, differences in speaker physiology, linguistic contexts, and imaging properties of the captured ultrasound data can significantly complicate the segmentation generalizability of the model. This highlights the importance of testing models on unseen datasets to evaluate their robustness and adaptability. As a result, generalizability across diverse datasets remains a critical hurdle. However, cross-dataset testing is often conducted on datasets with similar imaging properties \cite{RN13, RN9} and the same spoken language of the dataset subjects \cite{RN9}. Some previous studies utilized all available datasets for training without performing true cross-dataset evaluations with an unseen test dataset \cite{RN14, RN15}. This practice restricts the applicability of such models to real-world scenarios, where ultrasound devices vary in imaging quality, resolution, probe characteristics, and the subjects' spoken language. 

Furthermore, the low signal-to-noise ratio (SNR) in ultrasound images, caused by speckle noise and weak tissue boundaries, makes accurate tongue contour extraction particularly difficult, especially given the tongue's soft tissue structure and its dynamic, complex movements during speech  \cite{RN16}.  Some recent studies \cite{RN18, RN17, RN9} reported the promising potential of convolutional neural networks (CNNs) in advancing ultrasound tongue imaging tracking, surpassing traditional techniques such as active contour models (e.g., EdgeTrak)\cite{RN19, RN15}. However, complex models can sometimes face challenges such as overfitting to a specific dataset's imaging conditions, potentially limiting their generalizability to datasets with differing properties, including brightness, noise levels, and pixel intensity distributions. Additionally, in many cases, segmenting the bright, high-intensity tongue region in ultrasound images may not require the full complexity of large deep learning architectures, including the baseline UNet \cite{RN20}, as the task itself is relatively well-defined. Such models frequently neglect the critical need for computational efficiency and real-time performance, which is essential for high-speed ultrasound devices, which are preferred to monitor tongue movements. 

To address these limitations, there is a pressing need for a lightweight, real-time segmentation model specifically optimized for ultrasound tongue image processing for real-world applications. Such a model must balance computational efficiency, robustness to noise, and segmentation accuracy while remaining adaptive to diverse imaging conditions and speaker populations. Therefore, in this paper, we propose UltraUNet, a lightweight variant of UNet \cite{RN20} specifically designed for real-time segmentation of tongue contours in ultrasound images. UltraUNet incorporates several innovations tailored to ultrasound tongue imaging challenges, improving accuracy and inference speed. First, the model integrates SE \cite{RN21} blocks selectively into deeper layers of the encoder to enhance channel-wise feature recalibration while maintaining computational efficiency. Second, it employs selective normalization, where Group Normalization is applied only in deeper encoder layers to ensure stability in small-batch training. To enhance robustness against noise artifacts, we incorporate a denoising module as one of the data augmentation techniques, mutually exclusive with other noise-imitating augmentations. This denoiser, implemented as a standard UNet, is pre-trained on unannotated ultrasound images with artificially added speckle noise. Additionally, we compared UltraUNet against 9 other models within a single dataset (Fig. 1). To further assess the model's robustness, we conduct detailed cross-dataset evaluations with one of the datasets used for training and 7 other datasets left unseen with differences in ultrasound devices, imaging conditions, speaker populations, and linguistic contexts.

\begin{figure}
    \centering
    \includegraphics[width=1\linewidth]{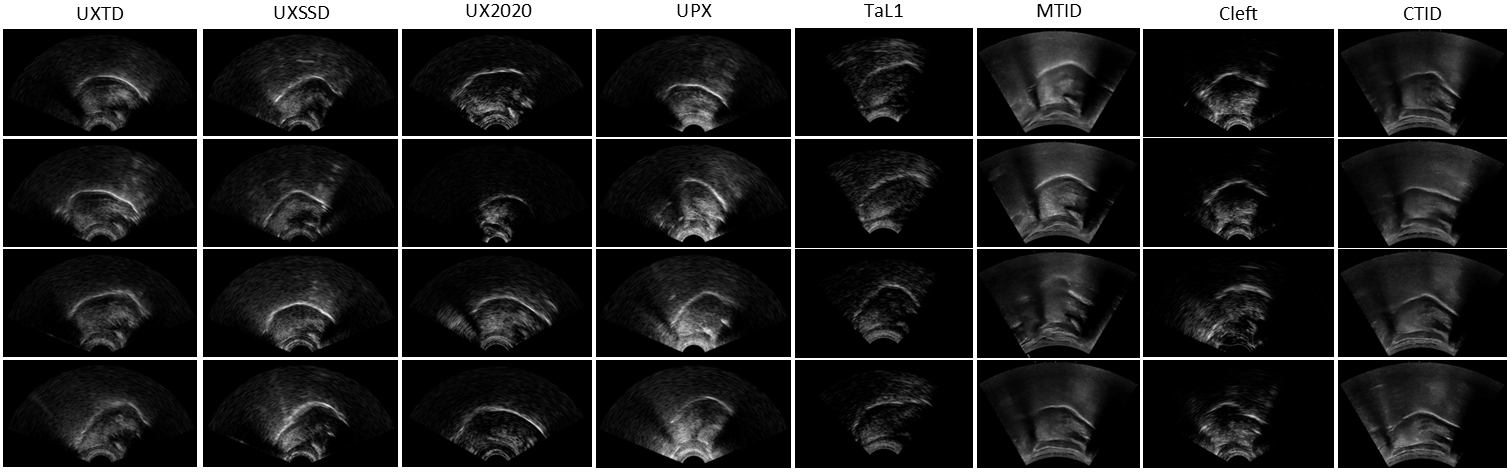}
    \caption{Sample images from the ultrasound datasets, including UXTD, UXSSD, UX2020, UPX, TaL1, MTID, Cleft, and CTID datasets, where each column has 4 randomly selected images from each of them.}
    \label{fig:placeholder}
    
    \includegraphics[width=0.5\linewidth]{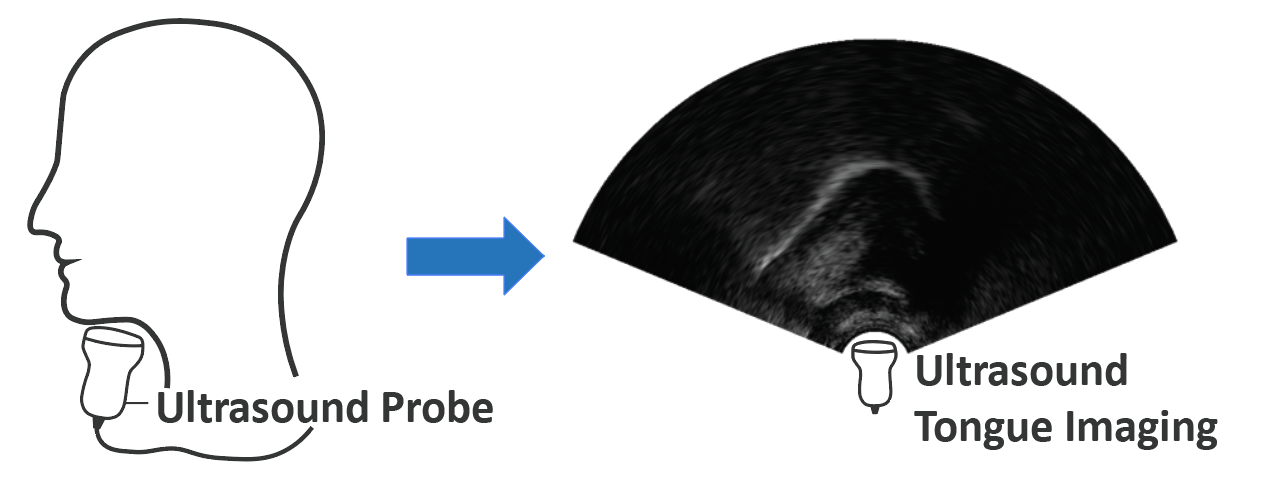}
    \caption{Ultrasound probe placement during data collection procedure of the MTID and CTID datasets with a single image example from MTID.}
    \label{fig:second_placeholder}
\end{figure}

\section{Related Works}
\subsection{Traditional Segmentation Methods}
In the tongue contour segmentation area, the main challenges include speckle noise and blur. There have been numerous advancements in adapting traditional contour tracking approaches, such as active contour models, including \cite{RN22, RN19}, where the user is required to define the contour only in the initial frame of the sequence, eliminating the need for further re-initialization. In addition, \cite{RN15} introduced automatic re-initialization mechanisms, such as Complex Wavelet Structural Similarity (CW-SSIM), which improved robustness by preventing contour drift. However, the method relied on manual initialization and exhibited slow processing (about 5 Hz) due to the computational complexity of multi-scale wavelet analysis, which evaluates structural similarities at fine-grained resolution. A proposed fully automated approach \cite{RN23} addressed these limitations by integrating phase symmetry filtering, skeletonization, and re-initialization strategies, achieving competitive accuracy but struggling with low-quality frames in impaired speakers due to difficulties faced while tuning their acquisition setup for subjects with physical limitations, as well as motor control difficulties \cite{RN23}. However, traditional methods remain limited by their reliance on handcrafted features and sensitivity to noise. Hence, deep learning offers a powerful alternative in the domain of tongue contour segmentation \cite{RN18}.

\subsection{CNN-Based Segmentation Methods}
The field of medical imaging has seen significant advancements in segmentation methods, particularly with encoder-decoder architectures like UNet \cite{RN20} and its variants. UNet’s skip connections and symmetrical design allow for effective integration of spatial and contextual features, while models like Attention UNet \cite{RN24} enhance this by focusing on relevant image regions using attention mechanisms. Similarly, other variants have explored integrating advanced feature refinement techniques, such as Squeeze-and-Excitation (SE) mechanisms \cite{RN25, RN26}, into the UNet framework, focusing on skip connections. However, channel recalibration in the top layers may not always be beneficial, as the features extracted in these stages are often superficial, and such additions can increase computational overhead. 

Thus, recent advancements in deep learning have seen the application of general-purpose architectures and task-specific models to the domain of tongue contour segmentation. For example, Dense UNet \cite{RN9} has demonstrated strong performance, excelling in cross-dataset generalization but at the cost of slower processing speed. Similarly, BowNet and wBowNet \cite{RN17} leveraged dilated convolutions to balance global and local feature extraction, achieving real-time performance (reported ~30–42 FPS) and higher precision compared to sUNet and sDeepLab, versions of UNet \cite{RN20} and DeepLab \cite{RN27}, respectively, simplified by the authors for fair comparison. Meanwhile, TongueNet \cite{RN13} redefined segmentation as a landmark detection task, achieving competitive accuracy while maintaining the fastest processing speed (reported 67 FPS). Recently, DAFT-Net \cite{RN14} introduced dual attention mechanisms, CBAM and Attention Gate, and entropy-based optimization for feature refinement, achieving high accuracy and real-time performance (reported ~29 FPS on 96x96 resolution images). However, its dual attention modules introduce computational overhead, limiting scalability to higher resolutions or ultra-high frame rates. Another noteworthy model achieving state-of-the-art performance in the tongue contour segmentation task is IrisNet \cite{RN28}, a model leveraging the innovative RetinaConv module inspired by peripheral vision. This architecture has been reported by \cite{RN28} and \cite{RN18} to deliver exceptional accuracy, surpassing UNet and wBowNet, making it a prominent benchmark in the field.

\section{Methodology}
\subsection{Datasets}
The datasets used in this study consist of annotated ultrasound tongue imaging data sourced from multiple collections, designed to capture diverse imaging characteristics and linguistic variations. The dataset list includes the Tongue and Lips Corpus (TaL1) \cite{RN29}, Ultrax Typically Developing Children Dataset (UXTD), Ultrax Speech Sound Disorders (UXSSD), Ultrax2020 (UX2020), Cleft Dataset\cite{RN6}, Mandarin Tongue Imaging Dataset and Cantonese Tongue Imaging Dataset (MTID and CTID, respectively). Only MTID and UXTD are used in both training and evaluation procedures, other datasets are exclusively used for testing purposes. Below, we provide detailed descriptions of the datasets used in this study:

\begin{figure}
    \centering
    \includegraphics[width=1\linewidth]{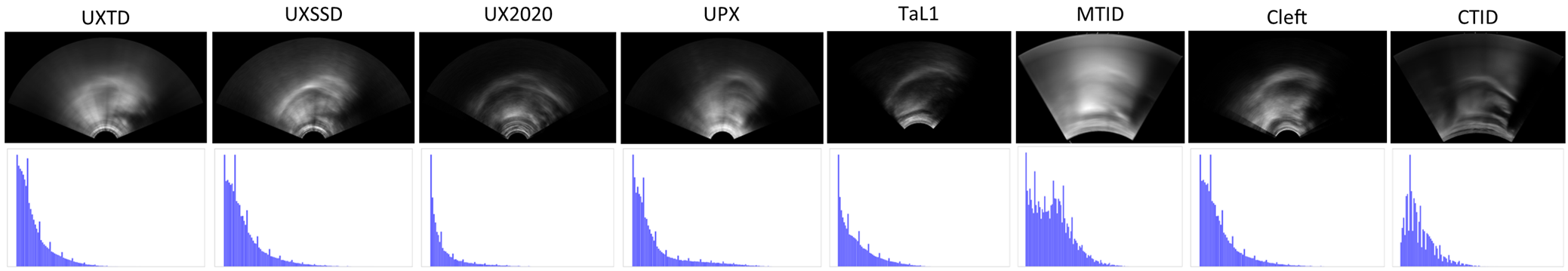}
    \caption{Average images generated by taking the average pixel intensity of each position (top row) and the intensity distribution plots (bottom row), both for each dataset.}
    \label{fig:placeholder}
\end{figure}

\subsubsection{Mandarin Tongue Imaging Dataset (MTID)}
The MTID dataset was collected at The Hong Kong Polytechnic University and consists of 1,611 annotated ultrasound tongue images acquired from 13 healthy Mandarin-speaking participants (7 males and 6 females; age range = 24–36 years). Participants were recruited from the campus and confirmed to have no history of neurological, speech, or hearing impairments. Ultrasound data acquisition was conducted using an ultrasound scanner and a convex probe (model with SC6-1 probe, SuperSonic Imagine Aixplorer, Aix-en-Provence, France). During the data collection process, participants were seated comfortably in an upright position to ensure consistent imaging conditions and natural tongue movement. A trained sonographer positioned the probe along the midsagittal plane beneath the chin, between the hyoid and mandible bones (Fig. 3) to capture high-resolution B-mode images of the tongue during a wide range of speech tasks, including sustained vowel phonation, dynamic utterances, and tasks with varied linguistic complexity such as vowels, single words, and consonant-vowel-consonant sequences, to comprehensively capture articulatory complexity. To ensure reproducibility and minimize variability, the participants were instructed to maintain a stable head position while following verbal instructions from the sonographer. The acquisition process adhered to standardized protocols for tongue ultrasound imaging, optimizing probe placement and acoustic coupling. All collected images were reviewed for quality assurance to exclude frames with excessive motion artifacts or incomplete tongue visualization. All experimental procedures were approved by the Human Subjects Ethics Sub-committee of The Hong Kong Polytechnic University (approval code: HSEARS20240327011, approved on 05-Apr-2024). This dataset exhibits a relatively broader intensity distribution (Fig. 4) compared to other datasets collected using different imaging hardware.

\subsubsection{Ultrax Typically Developing Children Dataset (UXTD)}
The UXTD dataset consists of 2,116 annotated ultrasound tongue images sampled from 58 typically developing English-speaking children (31 females, 27 males), aged 5–12 years. The data were collected using a different ultrasound imaging system by \cite{RN6}, resulting in unique imaging characteristics such as distinct contrast, resolution, and pixel intensity distributions (Fig. 4). 

\subsubsection{Additional Datasets for Testing}
To evaluate generalization of different models across different populations and imaging conditions, we included six additional datasets, each having 100-120 annotated images and representing distinct linguistic, clinical, and demographic characteristics:

\begin{itemize}
    \item \textbf{UltraPhonix (UPX)}: contains ultrasound tongue imaging data from 20 children (16 males, 4 females), aged 6–13 years, with speech sound disorders \cite{RN6}.
    \item \textbf{Ultrax 2020 Dataset (UX2020)}: consists of ultrasound and audio data from 37 English-speaking children (11 females, 26 males), aged 5–12 years, with speech sound disorders. The dataset was collected in hospital environments by speech and language therapists \cite{RN6}.
    \item \textbf{Ultrax Speech Sound Disorders Dataset (UXSSD)}: comprises data from 8 children (2 females, 6 males), aged 5–10 years, diagnosed with speech sound disorders \cite{RN6}.
    \item \textbf{Cleft Dataset}: includes ultrasound and audio recordings from 29 English-speaking children (11 females, 18 males), aged 7–11 years, with cleft lip and palate \cite{RN11}.
    \item \textbf{Tongue and Lips Corpus (TaL1)}: is a single-speaker dataset featuring ultrasound tongue images and synchronized video data of the lips from a professional male voice talent. The dataset spans six recording sessions, providing high-quality imaging data \cite{RN29}.
    \item \textbf{Cantonese Tongue Imaging Dataset (CTID)}: is collected under similar conditions to MTID, includes ultrasound tongue images from Cantonese-speaking participants. The imaging setup, participant recruitment criteria, and linguistic tasks mirrored those of the MTID dataset. Similarly to the MTID dataset collection, all our experimental procedures were approved by the Human Subjects Ethics Sub-committee of The Hong Kong Polytechnic University (ethics approval code: HSEARS20240306010, approved on 06-Mar-2024).
\end{itemize}

All images were annotated under the supervision of a trained sonographer using a custom-built annotation tool, providing pixel-wise labels that serve as ground truth for segmentation. Random sample images from all the datasets listed above are shown in Fig. 2. The intensity distributions and average images of the datasets are presented in Fig. 4, where the average image for each dataset is computed by taking the mean pixel intensity at each spatial location across all images within that dataset.

\subsection{Augmentations and Preprocessing}
As the test datasets possess varying imaging conditions, the primary purpose of augmentation techniques was to create different noise levels in addition to creating new artificial imaging samples. For this purpose, we incorporated a denoising augmentation strategy into the training pipeline. A separate UNet-based denoising model was trained on more than 40,000 unannotated ultrasound images collected from the UXTD and MTID datasets. The denoising model was optimized using the Adam optimizer with a learning rate of 0.001 and was trained for 10 epochs. The dataset was split into 80\% training and 20\% validation subsets, and the model that achieved the best performance on the validation set was selected for augmentation purposes. The denoising model learned to reconstruct clean ultrasound images from artificially corrupted inputs, simulating a wide range of signal-to-noise ratio (SNR) levels. During the main training process, this denoising model was employed to generate augmented training samples with varying noise properties, effectively expanding the diversity of the training set and improving the generalization capability of the model.  
In addition to denoising augmentation, other augmentation strategies included random horizontal flips to simulate variations in tongue orientation, speckle noise to inject artificial noise, and point spread function (PSF) to simulate the optical system blurring effect \cite{RN30}. It is important to note that the denoising augmentation was mutually exclusive with the speckle noise and PSF blur in the training pipeline to expose the models to varying noise conditions. Speckle noise and PSF artificially increase the noise levels to mimic real-world imaging artifacts, whereas the denoising model generates sample images with a lower noise level to enhance robustness in imaging conditions with less noise, and additionally by generating new artificial training samples. Preprocessing was employed to ensure effective training. All images were resized to 224x224 pixels to standardize input dimensions. Pixel intensity values were normalized to a [0, 1] range to ensure consistent input scaling. Full histogram matching \cite{RN31} was used to normalize pixel intensity distributions across datasets, reducing variability in imaging properties such as brightness levels. This ensures that the model is exposed to consistent intensity patterns during both training and testing, enabling it to generalize better to unseen datasets with differing imaging characteristics.

\begin{figure}
    \centering
    \includegraphics[width=1\linewidth]{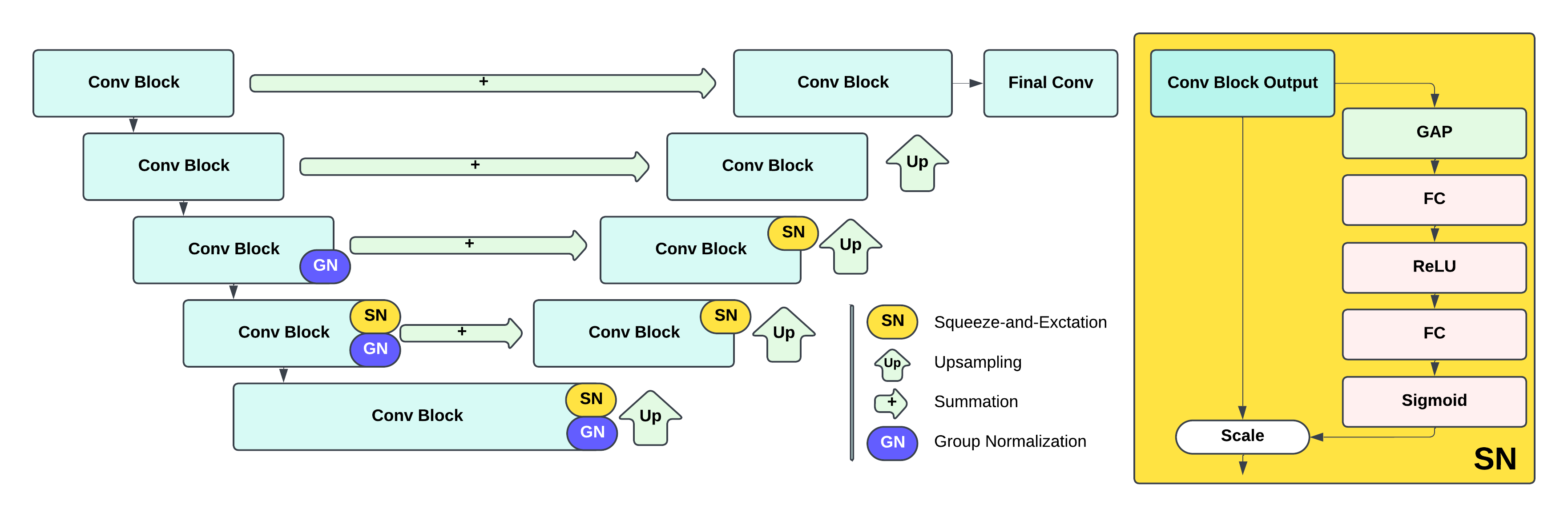}
    \caption{UltraUNet architecture for real-time ultrasound tongue segmentation. The encoder consists of sequential UltraUNet blocks with increasing channel depth, using Group Normalization and SE blocks selectively in deeper layers. Skip connections are implemented via summation. The decoder path includes upsampling operations and SE refinement in deeper layers. The final segmentation mask is produced using a 1×1 convolution. The right panel details the SE block, which employs Global Average Pooling (GAP) and fully connected (FC) layers with ReLU and sigmoid activations for channel-wise feature scaling.}
    \label{fig:placeholder}
\end{figure}

\subsection{Model Architecture}
The architecture of UltraUNet (Fig. 5) is a lightweight encoder-decoder design derived from the UNet model \cite{RN20}, incorporating a unique combination of modifications tailored to the challenges of ultrasound tongue imaging. The encoder path consists of five convolutional blocks with a kernel size of 3, each reducing spatial dimensions while increasing the number of feature channels. The number of channels starts at 24 and doubles at each subsequent stage, maintaining a lightweight design. SE blocks, introduced by \cite{RN21}, were proven to be effective in preliminary network design experiments and were selectively added to the deeper layers of the network to recalibrate channel-wise features, enhancing feature discrimination while incurring minimal computational overhead. The primary purpose of the module in our network design is to integrate channel-wise attention to suppress noise channels only in deeper layers of UltraUNet, where more abstract features are extracted. Group Normalization was applied in deeper encoder layers to stabilize training, particularly in small-batch settings \cite{RN32}, a common scenario in medical imaging segmentation tasks.

The decoder path mirrors the encoder, progressively reconstructing the segmentation mask through up-convolution layers. Summation-based skip connections, as opposed to concatenation, were employed to reduce memory and computational overhead while retaining essential spatial information. SE refinement was applied in deeper layers to refine feature maps further, ensuring accurate contour delineation. Additionally, the decoder path omits normalization layers to prioritize inference speed, a crucial requirement for real-time applications.

\subsection{Training and Evaluation}
UltraUNet and other models were trained using the Adam optimizer \cite{RN33} with a learning rate of 0.001 with a LambdaLR scheduler based on a custom polynomial decay function (Eq. 1), and training was conducted with a batch size of 3 for 50 epochs. 
\begin{equation}
\textit{factor} = \left( \frac{1 - e}{e_{\text{max}}} \right)^{0.9}
\end{equation}
where \(e\) is the current epoch number, and \(e_{max}\) is the maximum number of epochs, which is 50, in our case.

We employed combined Dice Focal loss (Eq. 2) as the loss function during the training and validation stages. Focal loss \cite{RN34}(Eq. 3) addresses the issue of class imbalance by reducing the relative loss contribution of well-classified examples and focusing more on hard-to-classify pixels. Dice loss is shown in (Eq. 4). The weights of Focal and Dice losses in the combined loss is 0.8 and 0.2, respectively.

\begin{equation}
L_{\text{comb}} = W_d \cdot L_d + W_f \cdot L_f
\end{equation}
where \(w_d\) is the weight for the Dice loss component, and \(w_f\) is the weight for the Focal loss component. 

\begin{equation}
L_{\text{Focal}} = -\alpha \cdot (1 - p_t)^\gamma \cdot \log(p_t)
\end{equation}

where \(\alpha\) is the weighting factor for class balance in Focal Loss, \(\alpha = 0.25\), and \(\gamma\) is the modulating factor in Focal Loss, \(\gamma = 2.0\).

\begin{equation}
L_{\text{Dice}} = 1 - \frac{2 \cdot \sum (p_t \cdot g_t) + \epsilon}{\sum p_t + \sum g_t + \epsilon}
\end{equation}

where \(p_t\) is the predicted probability for the target class, \(g_t\) is the ground truth label, and \(\epsilon\) is the smoothing term to avoid division by zero, \(\epsilon={10}^{-6}\)
Early stopping, with a patience value of 10, based on validation loss, was employed to prevent overfitting. All models were trained in PyTorch with CUDA on a PC with RTX3060 and 11th Gen Intel Core i5-11400, To evaluate generalization, cross-dataset evaluations were performed by training and validating the model on one dataset and testing it on the other datasets, which have different imaging properties. To minimize the effects of random initialization and other sources of variability, each model was trained three times, and the average results across these runs were reported.

The evaluation metrics used to measure the performance of UltraUNet included the Mean Sum Distance (MSD) (Eq. 5) and Dice Score \(Dice=\ 1-L_{Dice}\). MSD quantifies the average pixel distance between predicted and ground truth contours, providing a fine-grained assessment of contour precision. Lower MSD values indicate higher precision, which is critical for clinical applications such as speech therapy. To calculate the MSD score value, the segmented areas were skeletonized, and the points of the skeleton were compared to the ground truth points from our annotations. In the MSD calculation, we only consider the largest connected component of the segmentation output. The Dice Score measures the overlap between the predicted and ground truth segmentation masks, serving as a standard metric for segmentation accuracy. These metrics collectively capture complementary aspects of segmentation performance, addressing both contour precision and overall mask accuracy.

\begin{equation}
MSD(U, V) = \frac{1}{2n} \left( \sum_{i=1}^n \min_j |v_i - u_j| + \sum_{i=1}^m \min_j |u_i - v_j| \right)
\end{equation}

where \(n\) and \(m\) are the total number of points target contour and generated sequences, respectively, and \(u_i\) and \(v_j\) are pairs of two-dimensions coordinates from two sequences of ground truth contour \(U=[u_1,u_2,\ldots,u_n]\)  and the predicted contour \(V=[v_1,v_2,\ldots,v_n]\)

To comprehensively evaluate the performance of UltraUNet, we compared it against a variety of established segmentation models. These models include both general-purpose architectures and lightweight designs optimized for real-time performance. Below is a brief description of the models used in our comparison study:
\begin{enumerate}
    \item \textbf{UNet:} A widely used encoder-decoder architecture designed for biomedical image segmentation, utilizing skip connections to combine low-level and high-level features \cite{RN20}.
    \item \textbf{Attention UNet:} An extension of UNet incorporating attention gates to suppress irrelevant regions in feature maps, improving focus on target structures \cite{RN24}.
    \item \textbf{ResNet-50 (FCN): }A fully convolutional network (FCN) based on the ResNet-50 backbone, leveraging residual connections for improved feature extraction and gradient flow\cite{RN35}.
    \item \textbf{IrisNet:} A model incorporating the RetinaConv module inspired by peripheral vision, reported to deliver state-of-the-art performance in tongue contour segmentation.
    \item \textbf{Mobile UNet:} A lightweight variation of UNet designed for mobile and real-time applications, with reduced computational complexity \cite{RN36}.
    \item \textbf{SegNet:} An encoder-decoder model with indexed pooling for efficient upsampling, often used for semantic segmentation tasks \cite{RN37}.
    \item \textbf{Squeeze UNet: }A SqueezeNet-inspired variant of UNet designed for efficiency, achieving a 12x reduction in model size and a 3.2x reduction in MAC operations compared to standard UNet \cite{RN38}.
    \item \textbf{Swin UNet:} A transformer-based segmentation model utilizing the Swin Transformer for capturing global and local context \cite{RN39}.
    \item \textbf{USEnet:} A UNet variant incorporating SE blocks for channel-wise feature recalibration. SE blocks are added after every encoder block (Enc-USEnet) or after both encoder and decoder blocks (Enc-Dec USEnet) \cite{RN26}.
\end{enumerate}

\begin{table}[ht]
   \caption{SINGLE-DATASET EVALUATION OF EACH MODEL ON THE UNSEEN PART OF THE SAME DATASET}
   \centering
   \begin{tabular}{lcccccc}
      \toprule
      & \multicolumn{2}{c}{MTID Evaluation} & \multicolumn{2}{c}{UXTD Evaluation} & \multicolumn{2}{c}{Average (Mean ± STD)} \\
      \cmidrule(lr){2-3} \cmidrule(lr){4-5} \cmidrule(lr){6-7}
      Model & MSD↓* & Dice↑* & MSD↓ & Dice↑ & MSD↓ & Dice↑ \\
      \midrule
      \textit{UltraUNet} & 1.018 & \textbf{0.861} & 0.969 & \textbf{0.848} & \textbf{0.993 ± 0.053} & \textbf{0.855 ± 0.008} \\
      UNet & 1.012 & 0.860 & 1.008 & 0.846 & 1.010 ± 0.039 & 0.853 ± 0.008 \\
      Attention UNet & 1.039 & 0.858 & 1.105 & 0.844 & 1.072 ± 0.083 & 0.851 ± 0.008 \\
      Mobile UNet & 2.309 & 0.716 & 1.108 & 0.832 & 1.708 ± 0.845 & 0.774 ± 0.085 \\
      Squeeze UNet & 1.149 & 0.838 & 1.046 & 0.838 & 1.097 ± 0.072 & 0.838 ± 0.002 \\
      USEnet & \textbf{0.986} & 0.860 & 1.001 & 0.845 & 0.994 ± 0.050 & 0.852 ± 0.009 \\
      IrisNet & 1.147 & 0.839 & \textbf{0.952} & 0.842 & 1.049 ± 0.130 & 0.840 ± 0.002 \\
      Swin UNet & 1.282 & 0.838 & 1.035 & 0.842 & 1.158 ± 0.142 & 0.840 ± 0.003 \\
      SegNet & 1.304 & 0.831 & 1.140 & 0.839 & 1.222 ± 0.117 & 0.835 ± 0.011 \\
      ResNet-50 (FCN) & 1.245 & 0.831 & 1.117 & 0.831 & 1.181 ± 0.092 & 0.831 ± 0.003 \\
      \bottomrule
   \end{tabular}
   \label{tab:single_dataset_evaluation}
   \small \textit{\\**Arrows indicate the direction of optimal performance: ↓ lower is better, ↑ higher is better. \textbf{Bold values} represent the best performance for each metric.}
\end{table}

\begin{figure}
    \centering
    \includegraphics[width=1\linewidth]{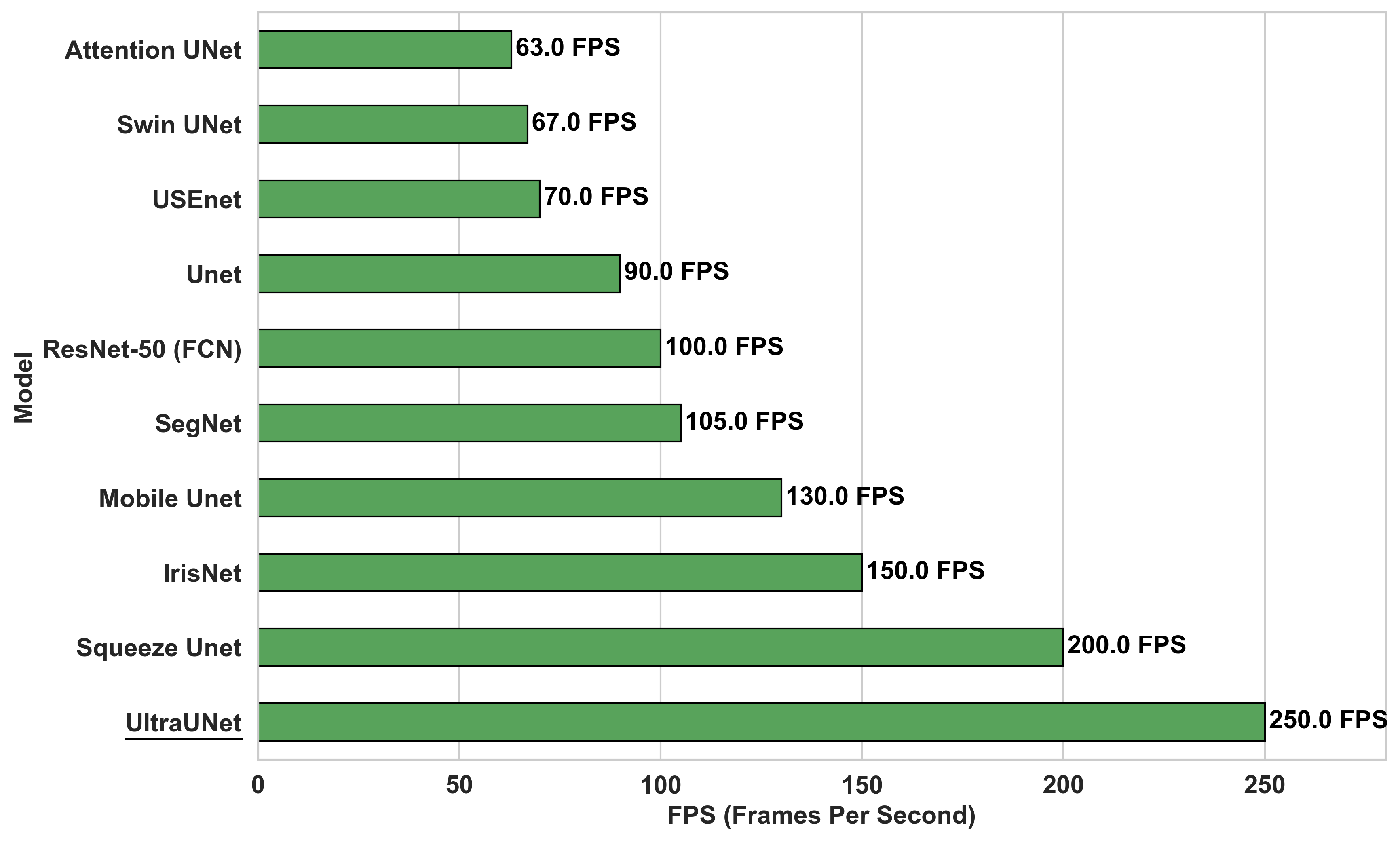}
    \caption{The inference speed, measured in Frames Per Second, for various segmentation models when processing input tensors of size (1, 1, 224, 224) continuously for 10 seconds.}
    \label{fig:placeholder}
\end{figure}

\begin{table}[ht]
   \centering
      \centering
      \caption{PROCESSING SPEED AND THE NUMBER OF PARAMETERS}
      \begin{tabular}{llll}
         \toprule    
         Model           & FPS & GFLOPs & Params  \\
         \midrule
         \textit{UltraUNet  }     & 250 & 6.005  & 4.454M   \\
         Unet            & 90  & 36.943 & 31.036M  \\
         Attention Unet  & 65  & 50.994 & 34.877M  \\
         Mobile Unet     & 130 & 1.106  & 13.341M  \\
         Squeeze Unet    & 200 & 1.601  & 2.502M   \\
         USEnet          & 70  & 50.187 & 34.744M  \\
         IrisNet         & 150 & 12.378 & 59.090M  \\
         Swin Unet       & 65  & 6.116  & 27.165M  \\
         SegNet          & 105 & 30.719 & 29.443M  \\
         ResNet-50 (FCN) & 100 & 26.526 & 32.94M   \\
         \bottomrule
      \end{tabular}
      \label{tab:processing_speed}
      \small \textit{\\All values are calculated based on the input size of an input tensor with the shape of (1, 1, 224, 224)}
\end{table}

\section{Experiments}
\subsection{Single-Dataset Evaluation}
Each dataset was divided into training, validation, and test subsets with an 80\%-10\%-10\% split, and the results reflect the models’ performance on the unseen part of the dataset. On the MTID dataset, UltraUNet achieved an MSD of 1.018 and a Dice Score of 0.861, slightly outperforming UNet (MSD = 1.012px, Dice = 0.860) and Attention UNet (MSD = 1.039px, Dice = 0.858) in terms of Dice Score, while maintaining a competitive MSD score. USEnet demonstrated the best MSD on MTID (MSD = 0.986px) while achieving a Dice Score of 0.860, comparable to UNet but slightly below UltraUNet. On the UXTD dataset, UltraUNet attained an MSD of 0.969 and a Dice Score of 0.848, showcasing robust segmentation accuracy. In comparison, UNet (MSD = 1.008px, Dice = 0.846) and Attention UNet (MSD = 1.105px, Dice = 0.844) performed slightly worse. IrisNet exhibited the best MSD on UXTD (MSD = 0.952px) but achieved a lower Dice Score of 0.842. When averaging results across both datasets, UltraUNet demonstrated a superior performance in both MSD and Dice (MSD=0.993px, Dice=0.855), underscoring its ability to maintain accurate contour precision and overlap accuracy. The average Dice Score vs FPS results for all models are illustrated in Fig. 1, and the inference speed comparison is illustrated in Fig. 6.  Each model was trained and evaluated three times to minimize the effect of random weight initialization and random augmentations, and the averaged results for each experiment and the average value for all six trials are listed in Table 1. Mobile UNet was excluded from the plot (Fig. 1) as it exhibited outlier behavior, with significantly poorer MSD and Dice Scores despite its lightweight design.

Table 2 shows the exceptional computational efficiency of UltraUNet, achieving a processing speed of 250 FPS with only 6.00G FLOPs and 4.45M learnable parameters, significantly outperforming UNet and Attention UNet in all metrics. UltraUNet operates approximately 2.78x faster than UNet and 3.57x faster than Attention UNet while maintaining a lightweight design with ~7x fewer parameters and ~7–8.5x lower FLOPs. While Mobile UNet has the lowest number of FLOPs, the inference speed is almost 2 times lower than UltraUNet’s. Even though Squeeze UNet has 2M fewer trainable parameters, UltraUNet achieves a 1.25x faster inference speed.

\begin{table}[ht]
   \caption{MEAN SUM DISTANCE (MSD) RESULTS FOR MODELS TRAINED ON MTID AND TESTED ON OTHER DATASETS IN EXPERIMENT 1}
   \centering
   \begin{tabular}{lcccccccc}
   \toprule
    \multicolumn{9}{c}{MSD↓* Score   Table (Trained on MTID)}                                \\
    \toprule
                    & Cleft  & CTID   & TaL1   & UPX    & UX2020 & UXSSD  & UXTD   & Average \\
    \textit{UltraUNet}       & 3.609  & 1.216  & 8.830  & 2.173  & \textbf{1.988}  & 1.691  & 4.151  & 3.380   \\
    Attention Unet  & 5.706  & 1.246  & 16.369 & 3.289  & 6.660  & 1.716  & 4.543  & 5.647   \\
    ResNet-50 (FCN) & \textbf{2.371}  & 1.259  & \textbf{5.083}  & 2.561  & 2.191  & \textbf{1.290}  & 4.350  & \textbf{2.729}   \\
    IrisNet         & 2.989  & 1.178  & 5.754  & 4.292  & 3.994  & 5.744  & 7.677  & 4.518   \\
    Mobile Unet     & 26.783 & 11.799 & 21.065 & 13.723 & 31.921 & 17.922 & 17.685 & 20.128  \\
    SegNet          & 3.353  & 1.241  & 12.414 & 3.223  & 3.235  & 2.863  & 5.296  & 4.518   \\
    Squeeze Unet    & 8.849  &\textbf{1.070}  & 7.994  & 2.495  & 4.452  & 2.039  & 4.853  & 4.536   \\
    Swin Unet       & 4.417  & 1.769  & 5.496  & 3.299  & 7.510  & 3.982  & 6.371  & 4.692   \\
    Unet            & 7.833  & 1.235  & 26.005 & 2.124  & 3.387  & 1.335  & \textbf{4.044}  & 6.566   \\
    USEnet          & 6.233  & 1.250  & 9.172  & \textbf{2.071}  & 3.234  & 1.335  & 4.398  & 3.956  \\
      \bottomrule
   \end{tabular}
   \small \textit{\\**Arrows indicate the direction of optimal performance: ↓ lower is better, ↑ higher is better. \textbf{Bold values} represent the best performance for each metric.}
\end{table}

\begin{table}[ht]
   \caption{DICE SCORE RESULTS FOR MODELS TRAINED ON MTID AND TESTED ON OTHER DATASETS IN EXPERIMENT 1}
   \centering
   \begin{tabular}{lcccccccc}
   \toprule
    \multicolumn{9}{c}{Dice↑* Score Table (Trained on MTID)}                                \\
    \toprule
                    & Cleft  & CTID   & TaL1   & UPX    & UX2020 & UXSSD  & UXTD   & Average \\
    \toprule
\textit{UltraUNet}    & 0.669 & 0.881 & 0.529 & 0.794 & \textbf{0.777}  & 0.816 & 0.669 & \textbf{0.734}   \\
Attention Unet  & 0.560 & 0.884 & 0.381 & 0.704 & 0.689  & 0.777 & 0.628 & 0.660   \\
ResNet-50 (FCN) & \textbf{0.697}& 0.872 & 0.542 & 0.699 & 0.702  & 0.768 & 0.621 & 0.700   \\
IrisNet         & 0.684 & 0.863 & 0.563 & 0.591 & 0.539  & 0.560 & 0.499 & 0.614   \\
Mobile Unet     & 0.005 & 0.380 & 0.131 & 0.267 & 0.112  & 0.237 & 0.179 & 0.187   \\
SegNet          & 0.588 & 0.879 & 0.528 & 0.735 & 0.580  & 0.770 & 0.634 & 0.673   \\
Squeeze Unet    & 0.261 & 0.868 & 0.318 & 0.755 & 0.684  & 0.780 & 0.640 & 0.615   \\
Swin Unet       & 0.651 & 0.874 & \textbf{0.614} & 0.700 & 0.594  & 0.683 & 0.567 & 0.669   \\
Unet            & 0.494 & \textbf{0.888} & 0.363 & 0.778 & 0.694  & 0.846 & 0.686 & 0.678   \\
USEnet          & 0.377 & 0.868 & 0.259 & \textbf{0.809} & 0.720  & \textbf{0.847} & \textbf{0.686} & 0.652   \\
\bottomrule
    \end{tabular}
   \small \textit{\\**Arrows indicate the direction of optimal performance: ↓ lower is better, ↑ higher is better. \textbf{Bold values} represent the best performance for each metric.}
\end{table}

\begin{table}[ht]
   \caption{MEAN SUM DISTANCE (MSD) RESULTS FOR MODELS TRAINED ON UXTD AND TESTED ON OTHER DATASETS IN EXPERIMENT 2}
   \centering
   \begin{tabular}{lcccccccc}
   \toprule
    \multicolumn{9}{c}{MSD↓* Score Table (Trained on UXTD)}                                \\
    \toprule
    & Cleft	& CTID &	MTID &	TaL1 &	UPX &	UX2020 &	UXSSD &	Average \\
    \toprule
    \textit{UltraUNet}       & \textbf{1.765} & 2.899 & \textbf{3.034}  & 3.705 & 2.370 & \textbf{1.934} & 1.544 & \textbf{2.464} \\
    Attention Unet  & 2.348 & 2.298 & 6.503  & 2.908 & 2.611 & 2.864 & 1.558 & 3.013 \\
    ResNet-50 (FCN) & 2.126 & 6.417 & 5.525  & \textbf{2.472} & \textbf{2.019} & 2.454 & \textbf{1.311} & 3.189 \\
    IrisNet         & 3.854 & 6.434 & 10.331 & 5.894 & 3.860 & 2.755 & 1.716 & 4.978 \\
    Mobile Unet     & 3.246 & 4.749 & 7.018  & 4.834 & 4.207 & 3.492 & 1.785 & 4.190 \\
    SegNet          & 3.186 & 3.245 & 5.131  & 3.131 & 5.520 & 6.151 & 1.726 & 4.013 \\
    Squeeze Unet    & 6.939 & \textbf{2.255} & 4.592  & 8.872 & 2.472 & 2.885 & 1.544 & 4.223 \\
    Swin Unet       & 2.342 & 2.435 & 5.190  & 5.606 & 2.694 & 2.640 & 1.794 & 3.243 \\
    Unet            & 2.762 & 4.703 & 7.548  & 4.758 & 3.008 & 3.330 & 1.594 & 3.958 \\
    USEnet          & 2.242 & 2.339 & 4.106  & 4.619 & 3.305 & 2.735 & 1.426 & 2.967 \\
\bottomrule
    \end{tabular}
   \small \textit{\\**Arrows indicate the direction of optimal performance: ↓ lower is better, ↑ higher is better. \textbf{Bold values} represent the best performance for each metric.}
\end{table}

\begin{table}[ht]
   \caption{DICE SCORE RESULTS FOR MODELS TRAINED ON UXTD AND TESTED ON OTHER DATASETS IN EXPERIMENT 2}
   \centering
   \begin{tabular}{lcccccccc}
   \toprule
    \multicolumn{9}{c}{Dice↑* Score Table (Trained on UXTD)}                                \\
    \toprule
    & Cleft	& CTID &	MTID &	TaL1 &	UPX &	UX2020 &	UXSSD &	Average \\
    \toprule
\textit{UltraUNet}       & \textbf{0.760} & 0.748 & \textbf{0.694} & 0.718 & \textbf{0.776} & \textbf{0.794} & 0.839 & \textbf{0.761} \\
Attention Unet  & 0.751 & 0.738 & 0.623 & 0.723 & 0.757 & 0.765 & 0.833 & 0.742 \\
ResNet-50 (FCN) & 0.747 & 0.674 & 0.634 & 0.732 & 0.774 & 0.774 & 0.834 & 0.739 \\
IrisNet         & 0.612 & 0.514 & 0.271 & 0.567 & 0.653 & 0.733 & 0.821 & 0.596 \\
Mobile Unet     & 0.676 & 0.660 & 0.559 & 0.640 & 0.655 & 0.709 & 0.793 & 0.670 \\
SegNet          & 0.684 & 0.747 & 0.640 & \textbf{0.733} & 0.586 & 0.661 & 0.824 & 0.696 \\
Squeeze Unet    & 0.583 & 0.772 & 0.646 & 0.499 & 0.743 & 0.751 & 0.830 & 0.689 \\
Swin Unet       & 0.714 & \textbf{0.778} & 0.612 & 0.625 & 0.735 & 0.746 & 0.812 & 0.717 \\
Unet            & 0.723 & 0.696 & 0.635 & 0.597 & 0.755 & 0.744 & 0.829 & 0.711 \\
USEnet          & 0.741 & 0.751 & 0.667 & 0.626 & 0.717 & 0.738 & \textbf{0.845} & 0.726 \\
\bottomrule
    \end{tabular}
   \small \textit{\\**Arrows indicate the direction of optimal performance: ↓ lower is better, ↑ higher is better. \textbf{Bold values} represent the best performance for each metric.}
\end{table}

\subsection{Cross-Dataset Evaluation}
To evaluate the generalizability of UltraUNet, we conducted cross-dataset experiments by training the model on one dataset and testing it on the other datasets with different imaging properties and/or subjects’ language or background. This subsection reports the Mean Sum Distance (MSD) and Dice Score results for two training configurations: training on the MTID dataset and testing on other datasets (Experiment 1) and training on the UXTD dataset and testing on the remaining datasets (Experiment 2). Comparisons with baseline models, including UNet, Attention UNet, and others, are included to highlight the performance advantages of UltraUNet. All the unseen datasets in both experiments are preprocessed using the histogram-matching technique described previously. Similarly to the previous experiment, each model was trained and evaluated three times to ensure fair comparison.

The MSD and Dice Score results, when the models were trained on MTID in Experiment 1, presented in Table 3 and Table 4, respectively, reveal that UltraUNet achieves the highest average Dice Score (0.734) while maintaining a competitive average MSD score (3.380), being the second-best performing model after a Fully Convolutional Network (FCN) with a ResNet-50 backbone in terms of MSD. In Experiment 2 (Table 5 and Table 6), with all the models trained on UXTD and tested on the other datasets, UltraUNet demonstrates a superior performance in both average MSD (2.464) and Dice Score (0.761), confirming its ability to generalize well to unseen data.
cross-dataset evaluations, UltraUNet demonstrated superior average performance compared to other models in terms of contour precision (MSD) and overlap accuracy (Dice). The integration of targeted augmentation strategies further improved the model’s ability to handle noise and variability, making it well-suited for real-world applications in ultrasound tongue imaging. The qualitative results, as illustrated in Figures 7 and 8, demonstrate the segmentation performance of all models. Each column corresponds to a different dataset, while each row represents a segmentation model. The ground truth contours (red lines) are compared against the predictions generated by the respective models. UltraUNet demonstrates closer alignment with the ground truth contours compared to other models, as evidenced by its smooth and accurate segmentation lines across datasets with varying noise levels and anatomical complexities. 

\begin{table}[ht]
\begin{minipage}[t]{0.48\textwidth} 
    \caption{AUGMENTATION RESULTS OF ULTRAUNET (TEST:MTID, TRAIN:UXTD)}
    \centering
    \begin{tabular}{ccc|cc}
        \toprule
        \multicolumn{3}{c|}{Augmentation Strategies} & \multicolumn{2}{c}{Test (MTID)} \\
        \cmidrule(lr){1-3} \cmidrule(lr){4-5}
        PSF & Speckle & Denoise & MSD↓ & Dice↑ \\
        \midrule
        \checkmark   &         &         & 5.029 & 0.608 \\
            & \checkmark       &         & 3.256 & 0.645 \\
            &         & \checkmark       & 3.313 & 0.650 \\
        \checkmark   & \checkmark       &         & 2.910 & 0.670 \\
        \checkmark   &         & \checkmark       & 3.367 & 0.662 \\
            & \checkmark       & \checkmark       & 3.290 & 0.666 \\
        \checkmark   & \checkmark       & \checkmark       & 2.711 & 0.689 \\
        \textbf{\checkmark} & \textbf{\checkmark} & \textbf{\checkmark} & \textbf{2.682} & \textbf{0.696} \\
        \bottomrule
    \end{tabular}
    \small \textit{Denoising is mutually exclusive with PSF and speckle noise augmentations.}
\end{minipage}
\hfill 
\begin{minipage}[t]{0.48\textwidth} 
    \caption{AUGMENTATION RESULTS OF ULTRAUNET (TEST: UXTD, TRAIN: MTID)}
    \centering
    \begin{tabular}{ccc|cc}
        \toprule
        \multicolumn{3}{c|}{Augmentation Strategies} & \multicolumn{2}{c}{Test (UXTD)} \\
        \cmidrule(lr){1-3} \cmidrule(lr){4-5}
        PSF & Speckle & Denoise & MSD↓ & Dice↑ \\
        \midrule
        \checkmark   &         &         & 6.288 & 0.621 \\
            & \checkmark       &         & 5.964 & 0.626 \\
            &         & \checkmark       & 5.919 & 0.609 \\
        \checkmark   & \checkmark       &         & 6.388 & 0.496 \\
        \checkmark   &         & \checkmark       & 5.793 & 0.606 \\
            & \checkmark       & \checkmark       & 5.774 & 0.613 \\
        \checkmark   & \checkmark       & \checkmark       & 6.700 & 0.544 \\
        \textbf{\checkmark} & \textbf{\checkmark} & \textbf{\checkmark} & \textbf{4.911} & \textbf{0.649} \\
        \bottomrule
    \end{tabular}
    \small \textit{Denoising is mutually exclusive with PSF and speckle noise augmentations.}
\end{minipage}
\end{table}

\subsection{Augmentations}
To identify the optimal augmentation strategies for UltraUNet, we conducted a systematic evaluation using the UXTD dataset as the test set and the MTID dataset as the training dataset, and vice versa. The goal was to study the impact of different augmentation strategies on the quality of segmentation results on the unseen dataset. In this experiment, to assess the impact of augmentation strategies fairly, the histogram matching technique was not utilized. The results, summarized in Table 7 and Table 8, show the impact of different combinations of augmentation techniques, including point spread function, speckle noise, denoising, and horizontal flipping. The best performance was achieved with the combination of PSF, denoising, speckle noise, and horizontal flipping, yielding an MSD of 2.68 and 4.91 on UXTD and MTID datasets, respectively, and a Dice Score of 0.70. The denoising augmentation was mutually exclusive with PSF blur and speckle noise. In our previous experiments, we used these augmentation strategies as the default augmentation pipeline for UltraUNet and the other models. 

\begin{figure}[ht]
\begin{minipage}[t]{0.48\textwidth} 
    \centering
    \includegraphics[width=\textwidth]{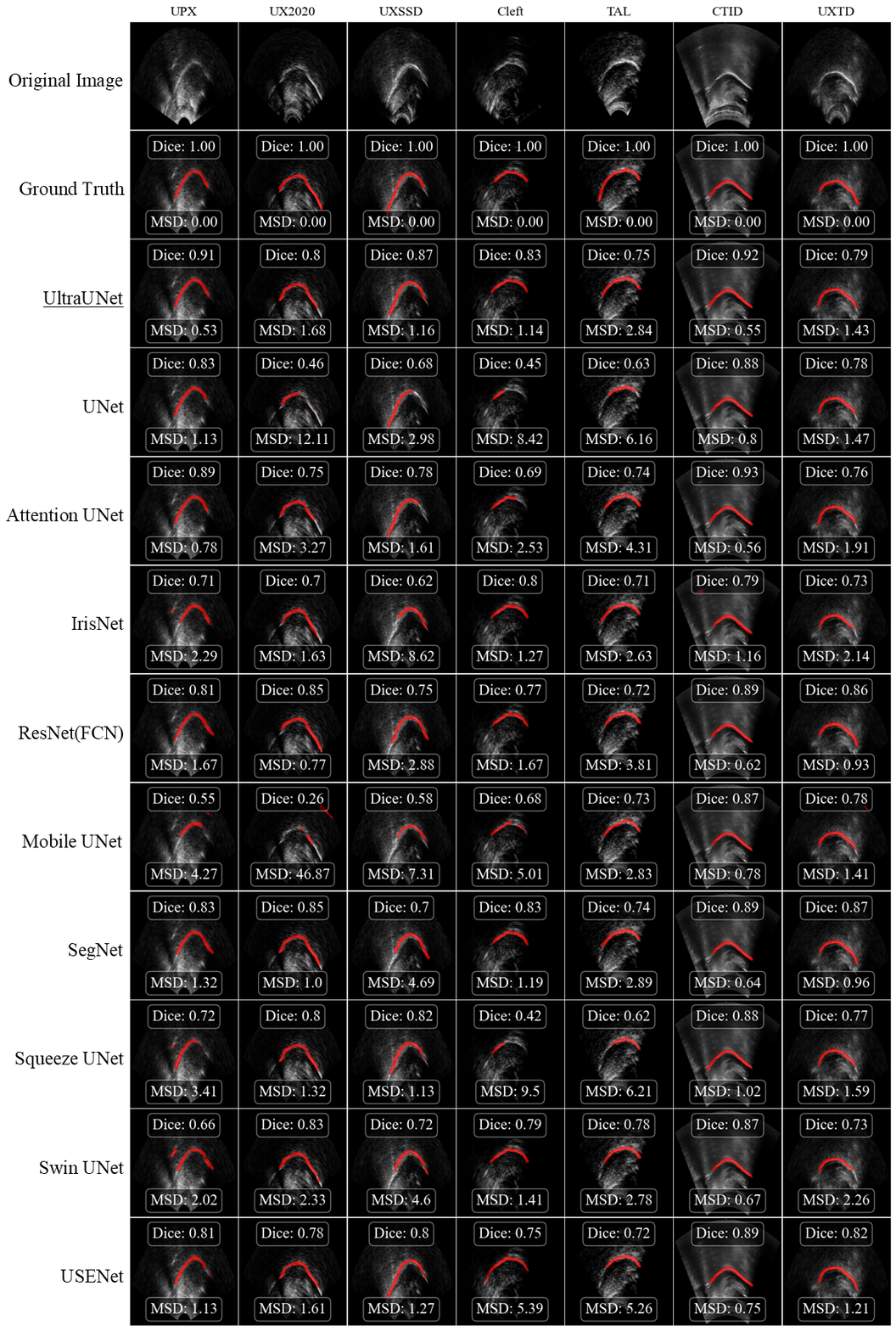} 
    \caption{The visualized segmentation results and extracted contours of the models trained on the MTID dataset in the cross-dataset evaluation experiment. Columns represent different datasets, from left to right: UPX, UX2020, UXSSD, Cleft, TaL, CTID, UXTD; and rows are as follows, from top to bottom: Original image, Ground truth, UltraUNet, UNet, Attention UNet, IrisNet, ResNet-50 (FCN), Mobile UNet, SegNet, Squeeze UNet, Swin UNet, USENet.}
\end{minipage}
\hfill 
\begin{minipage}[t]{0.48\textwidth} 
    \centering
    \includegraphics[width=\textwidth]{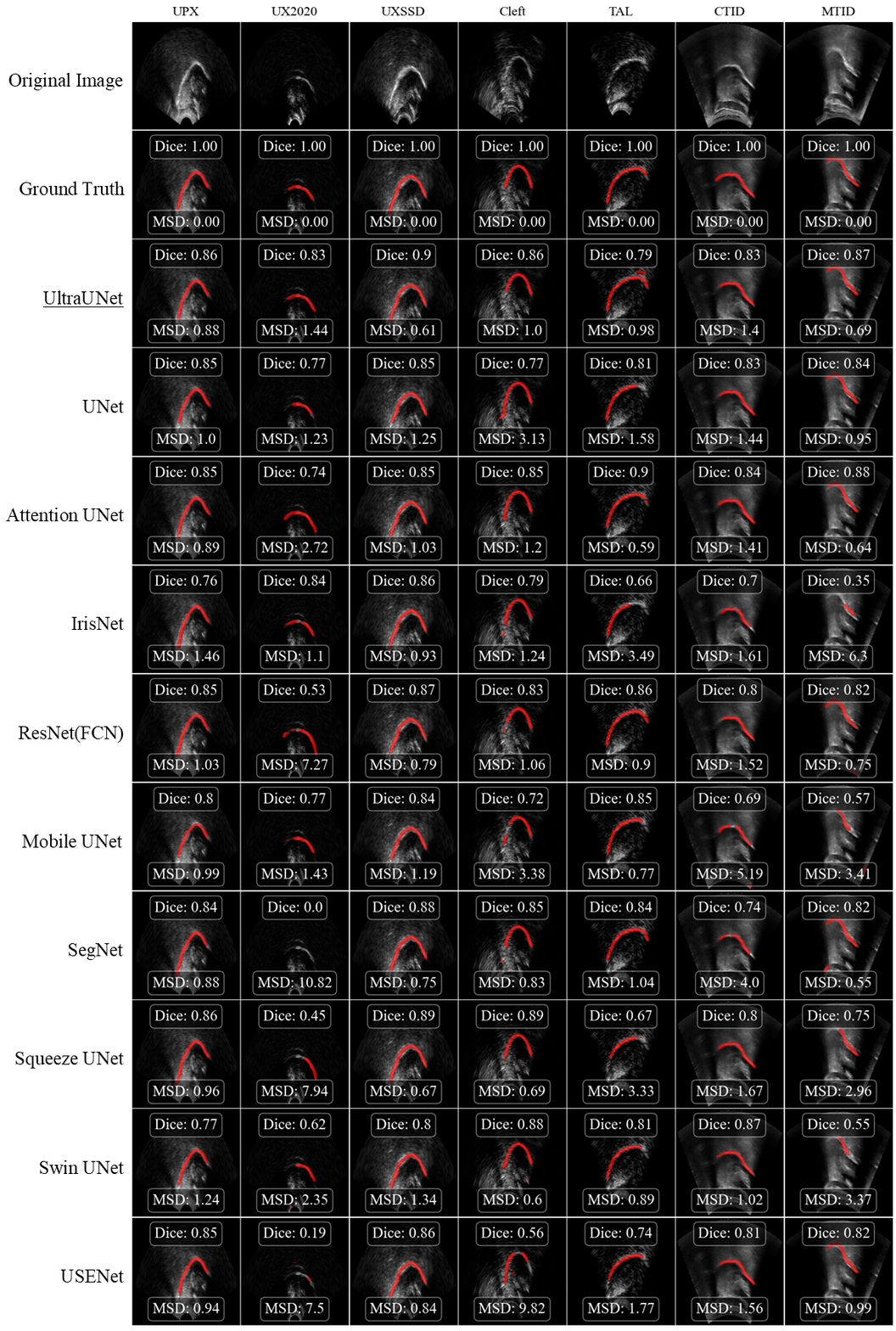} 
    \caption{The visualized segmentation results and extracted contours of the models trained on the UXTD dataset in the cross-dataset evaluation experiment. Columns represent different datasets, from left to right: UPX, UX2020, UXSSD, Cleft, TaL, CTID, MTID; and rows are as follows, from top to bottom: Original image, Ground truth, UltraUNet, UNet, Attention UNet, IrisNet, ResNet-50 (FCN), Mobile UNet, SegNet, Squeeze UNet, Swin UNet, USENet.}
\end{minipage}
\end{figure}

\section{Discussion and Conclusion}
\subsection{Discussion}
UltraUNet's efficiency is based on selecting highly optimized and effective operations, reducing computational overhead while preserving high segmentation accuracy. Therefore, several ineffective tongue contour segmentation methods that are traditionally used in general biomedical image segmentation were excluded. Both Batch Normalization and Group Normalization were observed to increase inference time significantly, especially in real-time applications. As a result, for stability in small batches \cite{RN32}, we selectively added Group Normalization to the deeper layers of the encoder where more abstract features are extracted. Additionally, our experiments showed that stabilizing high-level abstract features improved performance, while normalizing low-level features in shallow layers was less effective and introduced unnecessary computational overhead. Similarly, 1x1 convolutions are widely used in different established CNN architectures, such as Attention UNet \cite{RN24}. These convolutions are effective for reducing dimensionality and channel recalibration. However, their computational efficiency can be limited by challenges in memory access patterns and parallelization, as highlighted in \cite{RN40}, which discusses similar bottlenecks in convolution operations and the need for optimized implementations to fully realize theoretical advantages. Therefore, in our work, we used the SE layers, which rather rely on the fully connected layers with optimized parallel computations, not requiring additional intervention. Concatenation operations in skip connections, another common practice in UNet-based architectures, also contributed to higher runtimes, as they require additional memory allocation and data reorganization. By replacing concatenations with summation and limiting normalization to deeper layers, UltraUNet achieves an optimal trade-off between computational efficiency and performance. However, further investigation is needed to confirm these findings with additional references and benchmarks.

Our experiments reveal that spatial attention mechanisms, as implemented in Attention UNet \cite{RN24} and Swin UNet \cite{RN39}, do not provide significant performance improvements for the segmentation task at hand (Sections 4.2 and 4.1). While spatial attention excels in complex segmentation scenarios requiring precise localization \cite{RN24}, it appears less critical for the relatively simpler task of tongue contour segmentation in ultrasound images. Instead, lightweight channel-wise attention (via SE blocks) proved more effective by focusing on channel recalibration to suppress noise and enhance relevant features. This aligns with prior research in SE block integration in deep learning models \cite{RN41, RN42}. Moreover, spatial attention led to poorer generalization in some cases during the cross-dataset testing, suggesting that its complexity might inadvertently overfit to dataset-specific patterns. This reinforces the suitability of UltraUNet’s lightweight design for tasks involving noisy and low-resolution ultrasound imaging.

It is worth noting that in our study, we reproduced the IrisNet architecture as described by \cite{RN28} but observed significant challenges in resolution fidelity due to excessive downsampling from six encoder and four decoder layers. As a result, the output image had 4 times lower resolution than the input image. To address this, we applied interpolation with a Gaussian blur to smooth predictions. Additionally, our implementation revealed that the trainable parameter count is higher than reported, highlighting ambiguities in their methodology and the need for further clarification.

While UltraUNet demonstrated strong performance across various experiments, a few observations warrant further exploration. The model's reliance on channel attention suggests that spatial attention may not be necessary for certain medical imaging tasks, though this conclusion is task-dependent and may not generalize to more complex segmentation problems. Additionally, the hardware-aware optimizations, though effective, may require further refinement to balance efficiency and flexibility across different hardware platforms. The augmentation strategies used in this study were effective but might benefit from further experimentation with advanced techniques, such as adversarial data augmentation or domain adaptation, to address dataset-specific biases. Furthermore, while UltraUNet's channel-wise attention via SE blocks proved sufficient for tongue contour segmentation, more complex speech therapy tasks, such as identifying tongue dynamics or articulatory patterns over time, may benefit from hybrid attention mechanisms. The integration of lightweight spatial attention with temporal attention modules could be beneficial to better capture the spatiotemporal nuances of tongue movement in ultrasound sequences.

\subsection{Conclusion}
In this study, we introduced UltraUNet, a lightweight real-time model specifically designed for tongue contour segmentation in ultrasound imaging. By integrating SE blocks, Group Normalization, and efficient skip connections, UltraUNet achieves a balance between computational efficiency and segmentation accuracy. Our results have demonstrated that UltraUNet consistently outperforms or matches the performance of established models in both single-dataset and cross-dataset evaluations, achieving a notable inference speed of 250 FPS with high precision (Dice = 0.855 and MSD = 0.993px in single-dataset evaluation). The integration of ultrasound-specific augmentation techniques, including denoising and PSF blur simulation, further enhanced UltraUNet's robustness to noise and variability across diverse datasets. Additionally, histogram matching during evaluation significantly improved the generalizability of all models, though UltraUNet demonstrated inherent robustness even without this step. These attributes make it uniquely positioned for real-time clinical applications, such as speech therapy and articulatory phonetics research.

Furthermore, our experiments have demonstrated that established large segmentation models like Attention UNet and UNet, while achieving exceptional results for complex segmentation tasks, introduce unnecessary computational complexity for relatively straightforward tasks like tongue contour segmentation, where the primary objective is to delineate a bright curve accurately. This highlights the importance of task-optimized architectures, such as UltraUNet, which achieve comparable or superior accuracy while significantly reducing computational overhead. The simplicity of the tongue segmentation task underscores the suitability of UltraUNet's lightweight design, which avoids overfitting within one dataset with the same imaging properties and maintains high inference speed and high accuracy regardless of imaging conditions, the language background of the subject, and noise levels, setting up a benchmark for real-time efficient applications in clinical and research settings. Future work may include exploring hybrid attention mechanisms for spatiotemporal dynamics, advanced augmentations for bias mitigation, and refining hardware optimizations for broader adaptability.


\bibliographystyle{unsrt}  
\bibliography{references}  






\end{document}